%% file: arxiv.tex
\documentclass{article}

\usepackage{arxiv}

\usepackage[utf8]{inputenc} 
\usepackage[T1]{fontenc}    
\usepackage[hidelinks]{hyperref}
\usepackage{url}            
\usepackage{booktabs}       
\usepackage{amsfonts}       
\usepackage{nicefrac}       
\usepackage{microtype}      
\usepackage{lipsum}		
\usepackage{graphicx}
\usepackage[numbers,sort&compress]{natbib}
\usepackage{doi}
\usepackage{enumitem}
\usepackage{amsmath}
\usepackage{algorithm}
\usepackage{algpseudocode} 
\usepackage{amsmath}    
\usepackage{bbm}

\usepackage{graphicx}
\usepackage{subcaption}

\title{Neural-Inspired Posterior Approximation (NIPA)}

\author{Babak Shahbaba\thanks{Corresponding author.} \\
	Department of Statistics\\
	UC Irvine\\
	\texttt{babaks@uci.edu} \\
	\And
    Zahra Moslemi \\
	Department of Statistics\\
	UC Irvine\\
	\texttt{zmoslemi@uci.edu} \\
}

\date{}

\renewcommand{\headeright}{}
\renewcommand{\undertitle}{}
\renewcommand{\shorttitle}{NIPA}

\hypersetup{
pdftitle={NIPA},
pdfsubject={Bayesian Inference},
pdfauthor={Shahbaba and Moslemi},
pdfkeywords={MCMC, Bayesian inference, Decision-making},
}

\begin{document}
\maketitle

\begin{abstract}
	\input{sections/abstract}
\end{abstract}


\input{sections/introduction}

\input{sections/literature}

\input{sections/preliminaries}

\input{sections/method}

\input{sections/experiments}

\input{sections/discussion}

\input{sections/acknowledgment}

\bibliographystyle{unsrtnat}

\bibliography{references/References} 

\end{document}

%% file: sections/abstract.tex
Humans learn efficiently from their environment by engaging multiple interacting neural systems that support distinct yet complementary forms of control, including model-based (goal-directed) planning, model-free (habitual) responding, and episodic memory–based learning. Model-based mechanisms compute prospective action values using an internal model of the environment, supporting flexible but computationally costly planning; model-free mechanisms cache value estimates and build heuristics that enable fast, efficient habitual responding; and memory-based mechanisms allow rapid adaptation from individual experience. In this work, we aim to elucidate the computational principles underlying this biological efficiency and translate them into a sampling algorithm for scalable Bayesian inference through effective exploration of posterior distribution. More specifically, our proposed algorithm comprises three components: a model-based module that uses the target distribution for guided but computationally slow sampling; a model-free module that uses previous samples to learn patterns in the parameter space, enabling fast, reflexive sampling without directly evaluating the expensive target distribution; and an episodic-control module that supports rapid sampling by recalling specific past events (i.e., samples). We show that this approach advances Bayesian methods and facilitates their application to large-scale statistical machine learning problems. In particular, we apply our proposed framework to Bayesian deep learning, with an emphasis on proper and principled  uncertainty quantification.

%% file: sections/introduction.tex
\section{Introduction}
Bayesian inference provides a principled framework for uncertainty quantification and exploration of complex parameter spaces, but exact posterior inference is often intractable. As a result, in practice we typically rely on approximate methods, most notably Markov chain Monte Carlo (MCMC) and variational Bayes (VB). MCMC constructs Markov chains that asymptotically converge to the target distribution and is therefore unbiased. 
Although simple sampling methods, such as the Metropolis algorithm \cite{metropolis59}, are often effective at exploring low-dimensional distributions, they can be very inefficient and computationally expensive for complex and high-dimensional models. 
To reduce the random walk behavior of Metropolis sampling, Hamiltonian Monte Carlo (HMC) \cite{duane87, neal11} uses gradient information to propose samples that are distant from the current state, but nevertheless have a high probability of acceptance. 
However, HMC and related methods face major scalability challenges, as gradient computations over massive datasets are computationally infeasible. Stochastic-gradient-based MCMC variants \cite{Welling11,Ahn12,chen14,Ding14} aim to alleviate this cost through data subsampling, but the resulting noise can degrade accuracy and hinder effective exploration \cite{betancourt15}.

Variational Bayes (VB) \cite{Jordan99,wainwright08} offers a complementary alternative by framing inference as an optimization problem in which a parameterized distribution is used to approximate the posterior, trading unbiasedness for speed and scalability. To this end, VB methods minimize the Kullback–Leibler (KL) divergence between the target distribution and a parametrized approximating distribution with respect to the variational parameters. 
This approach has been further popularized and advanced through the development of efficient gradient-based optimization methods for variational objectives in deep learning \cite{kingma2014VAE, rezende2014stochastic, blundell2015weight}.
Compared to MCMC methods, VB is typically faster but introduces approximation bias. In addition, VB methods, particularly commonly used mean-field formulations, tend to underestimate posterior uncertainty \cite{Blei17}.

The contrast between these two common approaches naturally motivates hybrid methods that aim to combine the strengths of both paradigms \cite{freitas01,zhang16c}. When designed appropriately, such methods can achieve a better balance between computational efficiency and approximation accuracy, effectively bridging the gap between MCMC and VB. The method proposed in this paper pursues this goal by developing a general framework inspired by how humans efficiently learn to guide decision making through multiple interacting neural systems that support complementary forms of control: 
model-based (goal-directed) planning, which computes prospective action values using an internal model of the environment and enables flexible but computationally expensive decisions; model-free (habitual) control, which uses value estimates to learn patterns, supporting fast, efficient habitual responding; and memory-based mechanism, which enables rapid adaptation from individual experience \cite{DawNivDayan2005, LengyelDayan2007}.

Motivated by this perspective, we propose a general framework, called Neural-Inspired Posterior Approximation (NIPA), that comprises three corresponding components: (1) a model-based module guided by the target distribution for high quality, but computationally slow sampling; (2) a model-free module that enables fast, reflexive sampling through learned patterns in the parameter space; and (3) an episodic-control module that supports rapid one-shot sampling by recalling past experiences (i.e., previous samples). Here, we implement a specific and simplified instantiation of this framework. In the brain, although there is evidence linking model-based, model-free, and episodic control to the prefrontal cortex, striatum, and hippocampus, respectively \cite{DawDayan2014, LengyelDayan2007}, the neural implementation of these systems is highly complex, relying on distributed networks spanning hippocampal, cortical, striatal, and prefrontal regions with substantial overlap across systems.

In the following sections, we first provide a brief overview of recent advances in computational Bayesian inference. We then present our method in detail and evaluate its performance through comparisons with several state-of-the-art algorithms. In particular, we focus on uncertainty quantification in deep learning models. Our proposed approach constitutes a general framework with many possible variations, depending on the specific sampling algorithm used in the model-based module and the learning algorithm employed in the model-free component. In addition, multiple strategies can be considered for determining which module is engaged at any given point. In the Discussion section, we outline several potential directions for further enhancing and extending the framework.

%% file: sections/literature.tex
\section{Related Work}

The literature on improving the efficiency of computational methods for Bayesian inference is quite extensive. (See for example, \cite{neal96a, neal93, PMCMC, mykland95, ProppWilson96, roberts97, gilks98, warnes01, freitas01, brockwell06, neal11, neal05, neal03, Beal03, MollerPettittBerthelsenReeves06, AndrieuMoulines06, KuriharaWellingVlassis06, cappe08, craiu09, wellingUAI09, GelfandMaatenChenWelling10, douc07, douc11, wellingTeh11, ZhaSut2011a, ahmadian11, girolami11, hoffman11, beskos11, calderhead12, shahbabaSplitHMC, ahn13, lanLMC15, lanICML14, lanAAAI14, ahnShahbabaWelling14, nemeth2021stochastic, alexos22, Coullon2023Efficient}.) As mentioned above, Bayesian inference typically requires the use of MCMC algorithms to simulate samples from intractable distributions. The overall efficiency of these algorithms is judged by how fast they converge to the target distribution and generate samples that are truly representative of the underlying distribution. Although the simple Metropolis algorithm \cite{metropolis59} is often effective at exploring low-dimensional distributions, it can be very inefficient for complex, high-dimensional distributions: successive states may exhibit high autocorrelation, due to the random
walk nature of the movement. As a result, because the effective sample size (i.e., number of samples after adjustment for autocorrelation) tends to be quite low, the convergence to the true distribution is very slow. Faster explorations can be obtained using, for example, Hamiltonian Monte Carlo (HMC) \cite{duane87, neal11}. HMC and its variants \cite{shahbabaSplitHMC, girolami11, lanLMC15} reduce the random walk behavior of Metropolis by taking $L$ steps of size $\epsilon$ guided by Hamiltonian dynamics, which uses gradient information, to propose states that are distant from the current state, but nevertheless have a high probability of acceptance. 

The main bottleneck of these MCMC algorithms, especially in high-dimensional or complex settings, is the repeated evaluation of the posterior density, its gradient, and other related geometric or statistical quantities, which typically involve the entire observed (or training) dataset. A natural question is how to derive effective approximation of these quantities that provides a good balance between accuracy and computation cost. One common approach is subsampling \cite{hoffmann10, wellingTeh11, chen2014, nemeth2021stochastic, alexos22, Coullon2023Efficient}, which restricts the computation to a subset of the observed data. This is based on the idea that big datasets contain a large amount of redundancy so the overall information can be retrieved from a small subset. However, in general applications, we cannot simply use random subsets since the amount of information we lose as a result of random sampling leads to non-ignorable loss of accuracy, which in turn has a substantial negative impact on computational efficiency \cite{betancourt15}. Therefore, in practice, finding good criteria and strategies for effective subsampling is challenging, although some recent methods aim to alleviate this issue \cite{Campbell2024}.

Alternative methods exploit smoothness or regularity in parameter space, a property that holds for most statistical models regardless of the observed data. As a result, one can expect accurate and compact approximations of the posterior density and its gradient to exist in parameter space. Learning-based methods can be used to discover such efficient representations, which sampling algorithms can then leverage to substantially reduce computational cost and improve scalability. To this end, computationally efficient surrogate functions can be constructed to replace otherwise expensive target functions \cite{liu01, rasmussen03, zhang15, meeds14, Marzouk2014, lan15, strathmann15, zhang16b, Li2019}.

More recently, the Calibration–Emulation–Sampling (CES) framework \cite{Cleary2021} has emerged as an effective approach for large-dimensional uncertainty quantification, particularly in inverse problems \cite{Lan2022Siam}, by calibrating models to obtain parameter–posterior evaluations, emulating the parameter-to-posterior map, and generating posterior samples via MCMC at substantially reduced computational cost. Standard CES \cite{Cleary2021} employs Gaussian Process (GP) emulators (surrogates), which typically suffer from cubic computational complexity and scalability limitations in high dimensions despite various mitigation strategies \cite{lan15,liu2020,bonilla2007,gardner2018,seeger2003}. To overcome this issue, \citet{Lan2022Siam} proposed to replace GP emulation with convolutional neural networks, enabling CES to scale from hundreds to thousands of dimensions. Following this line of work, \citet{moslemi2024scaling} proposed a CES-based method for Bayesian inference in neural networks that employs deep (non-Bayesian) neural networks as emulators, leveraging their superior computational efficiency and flexibility for approximating high-dimensional functions.

Another class of algorithms aims to combine MCMC and VB methods to leverage the strengths of both approaches. An early attempt in this direction was proposed by \citet{freitas01}, who used a variational approximation as the proposal distribution in a block Metropolis–Hastings (MH) algorithm to rapidly identify high-probability regions and thereby accelerate convergence. Integration of variational inference and MCMC was also explored by \citet{Salimans15}, where one or more MCMC steps are embedded within the variational approximation. \citet{zhang16c}, on the other hand, proposed a framework that incorporates variational approximations into Hamiltonian Monte Carlo (HMC) to reduce the expensive computations required during sampling. They showed that additional flexibility introduced by MCMC steps enables a rich family of distributions that more accurately approximate the exact posterior compared to commonly-used VB methods.

In this paper, we propose a general framework that unifies all these methods as special cases. Rather than relying on a single exact or approximate method, our framework explores the parameter space by incorporating both approaches as distinct modules and dynamically selecting between them through a mechanism inspired by how humans engage different neural systems when exploring their environment and making decisions.

%% file: sections/preliminaries.tex
\section{Preliminaries}

While our proposed framework can be used for Bayesian inference in general, we discuss its details below in the context of Bayesian deep learning, where computational complexity is typically prohibitive. Standard neural networks (NNs) consist of an input layer $\boldsymbol{l}_0=\boldsymbol{X}$, a sequence of hidden layers $\boldsymbol{l}_l$ for $l=1,\ldots,m-1$, and an output layer $\boldsymbol{l}_m=\boldsymbol{Y}$. Each layer applies an affine transformation followed by a nonlinear activation function $g_l$ \citep{jospin2022}:
\begin{equation*}\label{eq:NN-structure}
\boldsymbol{l}_l = g_l\left(\boldsymbol{W}_l \boldsymbol{l}_{l-1} + \boldsymbol{b}_l\right), \quad l=1,\ldots,m-1.
\end{equation*}
The network parameters are denoted by $\boldsymbol{\theta}=(\boldsymbol{W},\boldsymbol{b})$, and a fixed architecture defines a function class indexed by $\boldsymbol{\theta}$. Given a training dataset $\{(x_n,y_n)\}_{n=1}^N$, learning involves estimating $\boldsymbol{\theta}$ to approximate the mapping $\boldsymbol{X}\rightarrow\boldsymbol{Y}$. The NN defines a forward operator $\mathcal{G}(\boldsymbol{\theta})$ such that
\begin{equation*}
\hat{y}_n = \mathcal{G}(x_n;\boldsymbol{\theta}), \qquad
y_n = \hat{y}_n + \varepsilon_n, \quad \varepsilon_n \sim \mathcal{N}(0,\Gamma),
\end{equation*}
where $\boldsymbol{Y}$ is continuous in regression problems or a continuous latent variable in classification.

Training typically involves solving
\begin{equation*}
\boldsymbol{\theta}^* = \arg\min_{\boldsymbol{\theta}\in\Theta}
L\!\left(\boldsymbol{Y} - \mathcal{G}(\boldsymbol{X};\boldsymbol{\theta})\right),
\end{equation*}
using stochastic gradient methods. Assuming Gaussian observation noise yields the negative log-likelihood loss
\begin{equation*}
\Phi(\boldsymbol{\theta};\boldsymbol{X},\boldsymbol{Y})
= \frac{1}{2}\left\|\boldsymbol{Y} - \mathcal{G}(\boldsymbol{X};\boldsymbol{\theta})\right\|_{\Gamma}^{2}.
\end{equation*}

Standard deep learning relies on point estimates of $\boldsymbol{\theta}$. This is computationally efficient but lacks proper uncertainty quantification (UQ) \citep{chuan2017,nixon2019}. To address this issue, Bayesian neural networks (BNNs) provide a principled UQ framework by placing a prior distribution $p(\boldsymbol{\theta})$ over parameters and performing Bayesian inference \citep{mackay1992,neal96}. The resulting posterior distribution is given by
\begin{equation*}
p(\boldsymbol{\theta}\mid X,Y) \propto p(Y\mid X,\boldsymbol{\theta})\,p(\boldsymbol{\theta}).
\end{equation*}

We assume a Gaussian observation model:
\begin{eqnarray*}
p(y \mid \theta) & = &
\prod_{i=1}^{N} \mathcal{N}\!\big(y_i \,;\, f_\theta(x_i), \sigma^2\big); \\
\ell(\theta) =
\log p(y\mid\theta)
& = & 
-\frac{1}{2\sigma^2}\sum_{i=1}^{N}\big(y_i-f_\theta(x_i)\big)^2 + C,
\end{eqnarray*}
with zero-mean Gaussian prior, $p(\theta)$. The posterior is therefore: $\pi(\theta)
\;\propto\;
\exp\!\big(\ell(\theta)\big)\,p(\theta).$

%% file: sections/method.tex
\section{Method}

We propose a general framework for posterior approximation inspired by human exploration, learning, and decision-making. Classical framework explain such behavior through the interaction of two computational systems \cite{DawDayan2014}: a model-based control supporting goal-directed behaviors and a model-free control supporting simpler behaviors, such as habits. Model-based mechanisms compute prospective action values using an internal model of the environment, supporting flexible but computationally costly planning, whereas model-free mechanisms use value estimates to enable fast and efficient responding. Despite their success, these accounts struggle to explain how decision makers adapt rapidly in environments characterized by sparse data, long temporal dependencies, and rich latent structure. The addition of a third system, episodic memory control, has been proposed to achieve rapid behavioral adaptation through memory retrieval and evaluation \cite{LengyelDayan2007}. Specifically, episodic memory control stores detailed representations of individual experiences and can bridge long-range dependencies \cite{GershmanDaw2017}. 

Building on these insights, we propose a Neural-Inspired Posterior Approximation (NIPA) framework for computationally efficient Bayesian inference. Our framework comprises the following three components:
\begin{itemize}[leftmargin=*]
    \item \textbf{Model-Based (MB):} Guided by the target distribution to support careful exploration of the parameter space, making it flexible but computationally slow.
    \item \textbf{Model-Free (MF):} Enables fast, reflexive sampling by building surrogate functions (analogous to heuristics and habits) based on past exploration of the parameter space.
    \item \textbf{Episodic Control (EC):} Supports rapid, one-shot sampling by recalling specific past events (i.e., samples).
\end{itemize}
Within this framework, the MB component (guided by the target distribution) explores the parameter space to generate high-quality, “well-thought-out” samples. These samples are then used by the MF components to learn and approximate expensive functions over the parameter space, and by the EC component to store memories that enable recall of past function evaluations for one-shot (i.e., made from a single recalled past evaluation without additional computation) accept–reject decisions. Note that many of the existing methods discussed earlier can be regarded as special cases of this general framework. Specifically, our method reduces to an exact sampling algorithm when relying solely on the MB component, and to a CES when relying solely on the MF component. The proposed framework integrates these two systems and further complements them with the EC module.

Below, we describe in detail a simple, specific implementation of this framework for posterior approximation. For the MB component, we use a standard HMC algorithm with exact evaluations of the log-posterior and its gradient. For the MF component, we employ a combination of an autoencoder and a feedforward deep neural network (DNN): the autoencoder reduces the dimensionality of the parameter space, and the DNN constructs a surrogate function in the resulting low-dimensional space. The EC component relies on retrieving previous values using a simple distance-based gating mechanism. We note that many alternative implementations are possible, some of which are discussed in the Discussion section.

\paragraph{Initial Pool Construction.}
We first run a standard stochastic-gradient Hamiltonian Monte Carlo (SGHMC) \cite{chen2014stochastic} sampler on the full-data posterior $\pi(\theta)$ using a conventional BNN parameterization.
From this SGHMC chain, we retain an initial set of $M_0$ posterior samples (typically, $M_0=100$) and compute their corresponding \emph{exact} log-posterior value, $\log \pi_{\mathrm{train}}(\theta^{(i)})$, using the full data (i.e., no minibatching): 
\[
\mathcal{P}_0 \;=\; \big\{(\theta^{(i)},\, \log \pi_{\mathrm{train}}(\theta^{(i)}))\big\}_{i=1}^{M_0},
\]
These samples are labeled as $\theta_{\text{MB}}$.

\paragraph{Surrogate Modeling.}
Using the initial samples $\{\theta^{(i)}\}_{i=1}^{M_0}$, we train an autoencoder to obtain a low-dimensional latent embedding of the parameter space,
\[
u \;=\; \mathrm{Encode}(\theta)\in\mathbb{R}^{d_u},
\qquad 
\theta \approx \mathrm{Decode}(u).
\]
We then embed all initial samples to form the training pairs
\[
\bigl(u^{(i)},\, \log \pi(\theta^{(i)})\bigr),
\qquad
u^{(i)}=\mathrm{Encode}(\theta^{(i)}),
\]
which are used to construct the surrogate model.
To this end, within the resulting low-dimensional space, we fit a DNN regression model $g(\cdot)$ to predict the \emph{log-posterior} from the latent code $u$:
\[
\widehat{\log \pi(\theta)} = g(u),
\quad \text{with } u=\mathrm{Encode}(\theta).
\]

\paragraph{Neural-Inspired Posterior Approximation (NIPA).}

We now describe the NIPA framework, which is composed of three components: MB, MF, and EC. 

At iteration $t$, we first generate a \emph{gating candidate} using a Gaussian random walk, $q_{\text{RW}}$:
\[
\tilde\theta=\theta+\sigma \varepsilon,\qquad \varepsilon\sim\mathcal N(0,I),
\]
and use its distance to the pool to determine which component should be activated. 
Let $\mathcal{P}$ denote the current pool (initialized as $\mathcal{P}_0$ and updated online).
We compute the distance from $\theta'$ to the pool in a standardized parameter space:
\begin{equation*}
d^{*}(\tilde\theta,\mathcal{P})
\;=\;
\min_{\theta_i \in \mathcal{P}}d(\tilde\theta,\mathcal{P})
\;=\;
\min_{\theta_i \in \mathcal{P}}
\left\|
\frac{\tilde\theta - m}{s}
-
\frac{\theta_i - m}{s}
\right\|_2,
\end{equation*}
where $m$ and $s$ are coordinatewise mean and standard deviation estimated from the current pool.
We then apply two thresholds, $t_1$ and $t_2$ with $t_1 < t_2$, to select one of the three components of NIPA: 

\begin{itemize}[leftmargin=*]
\item \emph{MB:}
If $d^{*}(\theta',\mathcal{P}) > t_2$, the absence of nearby samples suggests that past memories and patterns identified by the surrogate function in the parameter space may be unreliable. In this case, the sampler continues exploring the parameter space using an exact-gradient HMC step based on the target distribution $\pi(\theta)$. If the HMC proposal, $\theta'$, is accepted, we label it as {$\theta_{\text{MB}}$} (exact evaluation) and add the corresponding pair, $(\theta_{\text{new}},\log\pi(\theta_{\text{new}}))$, to the pool. 

\item \emph{MF:}
If $t_1 < d^{*}(\theta',\mathcal{P}) \le t_2$, we are in a region where there are no comparable memory episodes, but there are sufficient samples for the surrogate function to be reliable. In this case, we treat the gating candidate as the proposal, $\theta' = \tilde{\theta}$, and apply the Metropolis acceptance probability, using the surrogate function to approximate the log-posterior density instead of evaluating it exactly:
    \[
    u'=\mathrm{Encode}(\theta'),\qquad \widehat{\log\pi(\theta')}:=g(u').
    \]
If accepted, $\theta'$ becomes the new state, we label it as {$\theta_{\text{MF}}$} (approximate evaluation), and add the corresponding pair, $(\theta',\widehat{\log\pi(\theta'))}$, to the pool. 

\item \emph{EC:}
If $d^{*}(\theta',\mathcal{P}) \le t_1$, it indicates strong memory of past experiences, and the proposal lies within the vicinity of a previous sample. In this regime, we again set $\theta' = \tilde{\theta}$, but avoid either evaluating its exact log-posterior or using the surrogate function to calculate the Metroplois acceptance probability.
Instead, we retrieve the nearest stored episode (sample) in the pool and use its cached log-posterior value as a proxy for the proposal. 
Accepted EC moves update the current state but are not inserted into $\mathcal{P}$. That is, we do not treat them as new memories. 
\end{itemize}

We update the surrogate function every $k$ (typically, 100) iterations. Algorithm \ref{alg:nipa} summarizes the exact implementation of NIPA. The resulting procedure combines fast, memorized or learned proxies with a careful and reliable decision-making mechanism. The EC component yields substantial acceleration by skipping log-posterior evaluations near previously explored states; the MF component provides inexpensive surrogate-based evaluations where sufficient samples exist to construct reliable heuristics; and the MB component safeguards exploration by periodically performing robust updates using the exact posterior distribution. This combination is designed to reduce computational cost while maintaining empirically stable posterior exploration. 

\begin{algorithm}[t!]
\caption{Neural-Inspired Posterior Approximation (NIPA)}
\label{alg:nipa}
\begin{algorithmic}[1]
\Require Target posterior $\pi(\theta)\propto \exp(\ell(\theta))p(\theta)$; thresholds $t_1<t_2$; initial pool size $M_0$; total iterations $T$; RW proposal $q_{\mathrm{RW}}(\cdot\mid\cdot)$; exact MB kernel $\mathcal{K}_{\text{MB}}$ (HMC with MH correction); encoder $\mathrm{Enc}$ and surrogate $g$; surrogate update interval $k$.

\vspace{0.2em}
\Statex \textbf{Definitions:}
\Statex \hspace{1.2em}\textbf{Pool:} $\mathcal{P}$ stores tuples $(\theta_i,\, L_i,\, o_i)$ with $o_i\in\{\text{MB},\text{MF}\}$;
$L_i=\log\pi(\theta_i)$ for $\text{MB}$ and $L_i=\widehat{\log\pi}(\theta_i)$ for $\text{MF}$.
\Statex \hspace{1.2em}\textbf{Standardized distance-to-pool:}
$d^*(\tilde\theta,\mathcal{P})=\min_{(\theta_i,\cdot,\cdot)\in\mathcal{P}}
\left\|\frac{\tilde\theta-m}{s}-\frac{\theta_i-m}{s}\right\|_2,
$
where $(m,s)$ are coordinatewise mean/std over $\{\theta_i:(\theta_i,\cdot,\cdot)\in\mathcal{P}\}$.
\vspace{0.2em}
\State Run an SGHMC chain targeting $\pi(\theta)$ and retain $M_0$ states $\{\theta^{(i)}\}_{i=1}^{M_0}$. \Comment{Initial pool construction}
\State $\mathcal{P}\leftarrow \emptyset$.
\For{$i=1$ to $M_0$}
    \State $L^{(i)} \leftarrow \log\pi(\theta^{(i)})$ \Comment{Exact, full log-posterior}
    \State Add $(\theta^{(i)},L^{(i)},\text{MB})$ to $\mathcal{P}$.
\EndFor
\State Train $\mathrm{Enc}$ on $\{\theta^{(i)}\}_{i=1}^{M_0}$ and $g$ on $\{(\mathrm{Enc}(\theta^{(i)}),\,\log\pi(\theta^{(i)}))\}_{i=1}^{M_0}$. \Comment{Build surrogate function}
\State Initialize $\theta \leftarrow \theta^{(M_0)}$.
\vspace{0.2em}
\For{$t=M_0+1$ to $T$} \Comment{Posterior approximation}

    \State \textbf{Gating candidate (RW):} draw $\tilde\theta \sim q_{\mathrm{RW}}(\cdot\mid\theta)$.
    \State Compute $d^*\leftarrow d^*(\tilde\theta,\mathcal{P})$ and retrieve nearest $(\theta_j,L_j,o_j)\in\mathcal{P}$.

    \If{$d^* > t_2$} \Comment{{MB}; Propose a new sample $\theta'$ via HMC}
        \State \textbf{Proposal generation (HMC):} sample $(p_0\sim\mathcal N(0,I))$ and run leapfrog steps to propose $\theta'$.
        \State \textbf{Accept probability:} $\alpha \leftarrow \min\{1,\exp(H(\theta,p_0)-H(\theta',p'))\}$.
        \State Draw $u\sim \mathrm{Unif}(0,1)$.
        \If{$u<\alpha$}
            \State $\theta \leftarrow \theta'$.
            \State $(\theta_{\text{MB}},L_{\text{MB}})\leftarrow(\theta',\log\pi(\theta'))$. \Comment{Exact evaluation}
            \State Add $(\theta_{\text{MB}},L_{\text{MB}},\text{MB})$ to $\mathcal{P}$.
    \EndIf

    \ElsIf{$t_1 < d^* \le t_2$} \Comment{{MF}; Treat $\tilde\theta$ as a proposal}
        \State $\theta' = \tilde{\theta}$    
        \State \textbf{Approximate function:} $L' \leftarrow \widehat{\log\pi}(\theta')=g(\mathrm{Enc}(\theta'))$
        \State \textbf{Reference:} $L_{\text{ref}} \leftarrow L_{\text{MF}}$
        \State \textbf{Accept probability:} $\alpha \leftarrow \min\{1,\exp(L'-L_{\text{ref}})\}$ \Comment{Metropolis update}
        \State Draw $u\sim \mathrm{Unif}(0,1)$.
        \If{$u<\alpha$}
            \State $\theta \leftarrow \theta'$.
            \State $(\theta_{\text{MF}},L_{\text{MF}})\leftarrow(\theta',L')$.
            \State Add $(\theta_{\text{MF}},L_{\text{MF}},\text{MF})$ to $\mathcal{P}$.
    \EndIf

    \Else \Comment{EC; Treat $\tilde\theta$ as a proposal}
    \State $\theta' = \tilde{\theta}$
        \State \textbf{Approximate function:} $L_{\text{proxy}} \leftarrow L_j$ \Comment{Retrieve cached value} 
        \State \textbf{Reference:} $
        L_{\text{ref}} \leftarrow
        \begin{cases}
            L_{\text{MB}} & \text{if } o_j=\text{MB},\\
            L_{\text{MF}} & \text{if } o_j=\text{MF}.
        \end{cases}$
        \State \textbf{Accept probability:} $\alpha \leftarrow \min\{1,\exp(L_{\text{proxy}}-L_{\text{ref}})\}$ \Comment{Metropolis update}
        \State Draw $u\sim \mathrm{Unif}(0,1)$; \textbf{if} $u<\alpha$ \textbf{then} $\theta\leftarrow \theta'$ \Comment{Do not add to $\mathcal{P}$}
    \EndIf
    \State Record $\theta^{(t)}\leftarrow\theta$.
    \If{$t \bmod k = 0$} \Comment{Update surrogate function}
    \State Refit surrogate model using current pool $\mathcal{P}$
\EndIf
\EndFor
\end{algorithmic}
\vspace{12pt}
\end{algorithm}

%% file: sections/experiments.tex
\section{Experiments}
 
\paragraph{Baseline Methods.} We evaluate our proposed approach against a set of baseline methods, including BNN models with the HMC sampler (BNN-HMC), SGHMC sampler (BNN-SGHMC), pCN sampler (BNN-pCN), Bayesian Lasso (BNN-LASSO) \cite{park08}, Variational Inference (BNN-VI) \cite{graves2011practical}, MC-Dropout (BNN-MCD) \cite{gal2016dropout}, Stochastic Weight Averaging-Gaussian (SWAG) \cite{izmailov2018, maddox2019simple}, and random network surrogate (BNN-RNS) \cite{Zhang2017a}. We also include results from a DNN, which does not provide uncertainty quantification but serves as a reference point. Moreover, we report results for the Ensemble Deep Learning method (DNN-Ensemble) \cite{lakshminarayanan2017}, which consists of multiple DNNs, each initialized with a different random seed. Although this approach allows for uncertainty quantification, it does not provide a probabilistic framework for analysis, a key advantage of BNN and its variants.

\begin{table}[t!]
    \caption{Comparison of NIPA with existing methods on regression problems. Computationally, NIPA is substantially more efficient than the BNN-HMC baseline, by factors of 8.27 on the synthetic data and 6.99 on the Year Prediction data, while achieving comparable RMSE and improved (higher) CP95 relative to the other methods.}
    \vspace{6pt}
    \label{tab:reg}
    \centering
    \begin{subtable}[t]{0.48\textwidth}
        \centering
        \includegraphics[width=\linewidth]{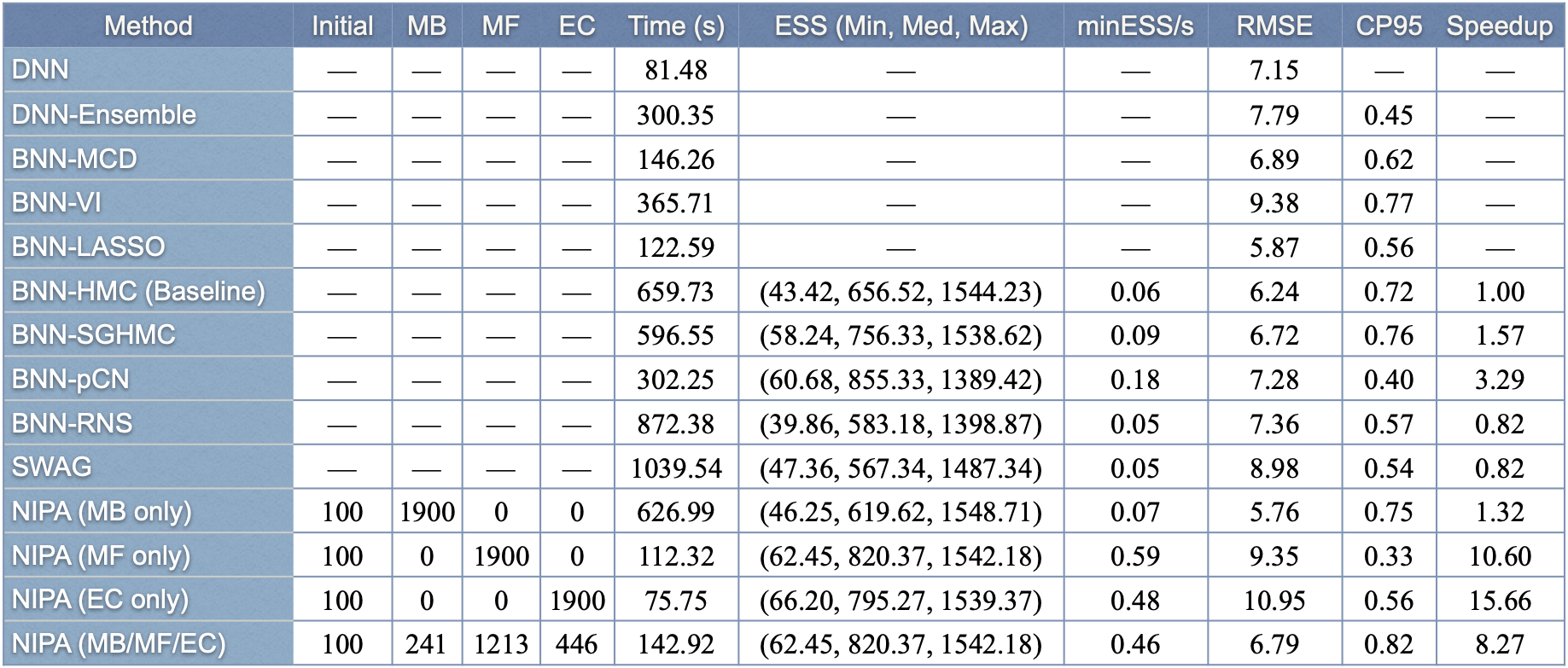}
        \vspace{-6pt}
        \caption{Synthetic Regression Problem.}
        \label{tab:nipa_reg}
    \end{subtable}
    \hspace{6pt}
    \begin{subtable}[t]{0.48\textwidth}
        \centering
        \includegraphics[width=\linewidth]{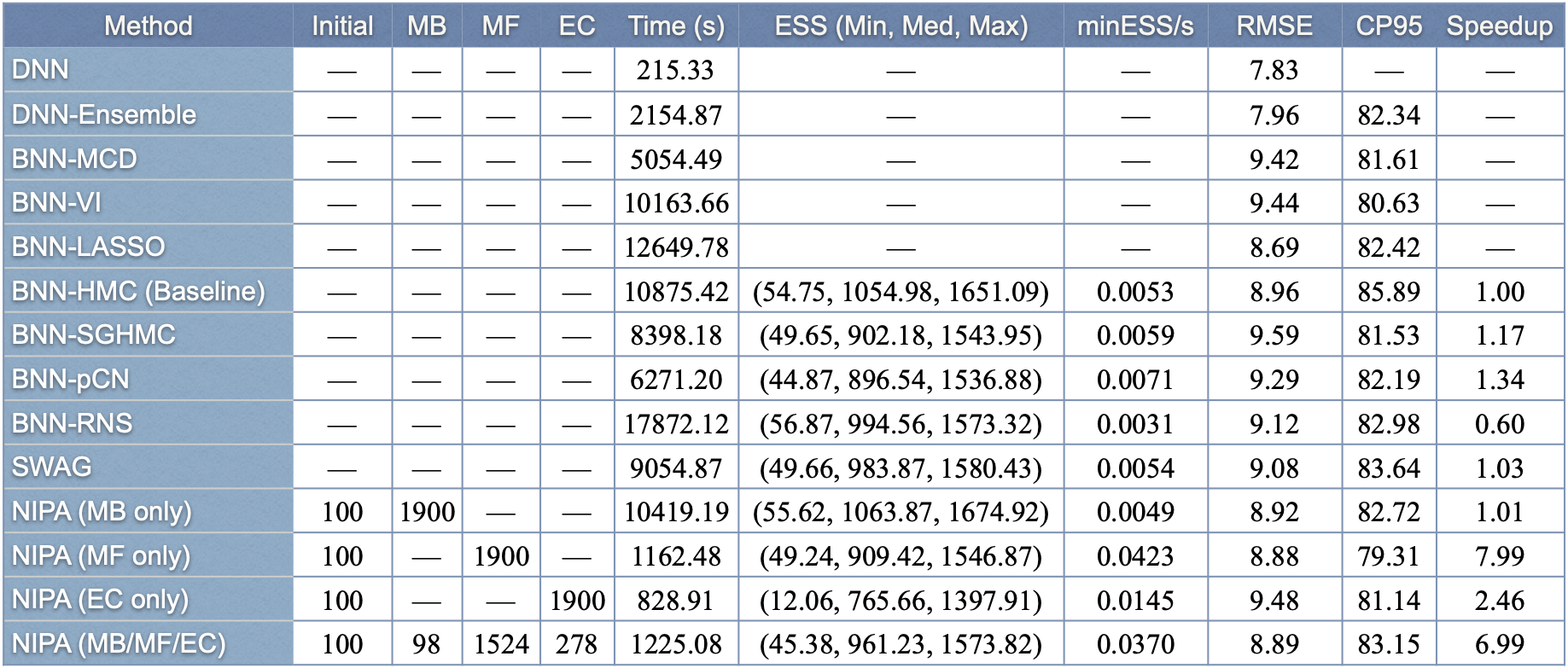}
        \vspace{-6pt}
        \caption{Year Prediction MSD.}
        \label{tab:nipa_msd}
    \end{subtable}
    
\end{table}

\paragraph{Metrics.}
For the regression benchmark, we report predictive performance using the root mean squared error (RMSE) on the held-out test set. We evaluate the quality of uncertainty quantification using the empirical coverage probability of nominal 95\% posterior predictive intervals (CP95), which is defined as the fraction of test targets $y_i$ that fall within the 95\% interval $[L_i, U_i]$,
\[
\mathrm{CP95} = \frac{1}{n}\sum_{i=1}^n \mathbbm{1}\{L_i \le y_i \le U_i\},
\]
Here, $[L_i, U_i]$ is constructed from the posterior predictive distribution at $x_i$. 

For the classification problems, we report test accuracy and evaluate uncertainty quality using the expected calibration error (ECE) \cite{naeini2015bayesian, guo17calibration}. 
For each test point $x_i$, we compute the posterior-mean predictive probabilities $\bar{p}(x_i)$ and assign a predicted class $\hat{y}_i = \arg\max_k \bar{p}_k(x_i)$ with confidence 
$c_i = \max_k \bar{p}_k(x_i).$
Test points are grouped into $B$ bins according to their confidence, and for each bin, $b$, we calculate the empirical accuracy $\mathrm{acc}(B_b)$ and average confidence $\mathrm{conf}(B_b)$. 
ECE is then defined as the weighted average absolute difference between these two quantities across bins:
\[
\mathrm{ECE} = \sum_{b=1}^B \frac{|B_b|}{n} \, \bigl| \mathrm{acc}(B_b) - \mathrm{conf}(B_b) \bigr|,
\]
where smaller values indicate better calibration.

In addition to predictive performance and uncertainty quantification, we assess computational efficiency and sampling quality using wall-clock runtime, $T$, and effective sample size (ESS) \cite{geyer92, geyer2011introduction}. Each sampling algorithm produces $S = 2000$ posterior samples $\{\theta^{(t)}\}_{t=1}^{S}$, where $\theta^{(t)} \in \mathbb{R}^{D}$ concatenates all weights and biases. We evaluate mixing at the level of individual scalar parameters: for each $j \in {1, \dots, D}$, we treat $\{\theta^{(t)}_j\}_{t=1}^{S}$ as a univariate Markov chain and compute its ESS. The resulting per-parameter ESS values are summarized by their minimum, median, and maximum. The minimum ESS (minESS) reflects the slowest-mixing direction. To combine sampling quality with computational cost, we report minESS per second, $\mathrm{minESS}/s$, which is then used to calculate the speedup of a sampling algorithm relative to the BNN-HMC baseline.

\paragraph{Regression.} We first construct a synthetic regression benchmark based on a two-hidden-layer feedforward neural network with ReLU activations, which maps randomly generated inputs to scalar outputs perturbed by additive Gaussian noise. Let $d$ denote the input feature dimension and $N$ the sample size. The network architecture consists of two hidden layers of widths $h_1$ and $h_2$. We concatenate all weights and biases into a single parameter vector $\theta \in \mathbb{R}^D$, where $D$ is determined by the layer dimensions. 

A design matrix $X \in \mathbb{R}^{N \times d}$ is sampled with i.i.d.\ uniform entries, where each row is treated as a predictor. A ground-truth parameter vector $\theta^\star$ with i.i.d elements is then sampled from a bounded uniform distribution, which defines the data-generating mechanism. The noiseless latent signal is obtained via a forward pass as $f(X;\theta^\star) \in \mathbb{R}^N$, and the observed responses are generated by adding Gaussian noise,
\[
y = f(X;\theta^\star) + \varepsilon, \qquad 
\varepsilon \sim \mathcal{N}\!\bigl(0,\operatorname{diag}(\sigma^2)\bigr),
\]
where the noise variance $\sigma^2$ is chosen to control the level of noise-to-signal ratio. The resulting dataset has 5,000 observations, 100 features, and 3,505 model parameters. 

We also evaluate our method based on the Year Prediction MSD dataset \cite{year_prediction_msd_203}, where the goal is to predict the release year of a song from audio features. Songs are mostly western, commercial tracks ranging from 1922 to 2011, with a peak in the year 2000s. This dataset has 515,345 observations, 90 features, and 309,761 model parameters.

Each dataset is partitioned into training and test sets, which are held fixed across all methods. The results on the test set are provided in Table \ref{tab:reg}. As we can see, NIPA is substantially more efficient than the BNN-HMC baseline, with comparable RMSE and higher CP95 than other methods.

\begin{table}[t!]
    \caption{Comparison of NIPA with existing methods on classification problems. Computationally, NIPA is substantially more efficient than the BNN-HMC baseline, by factors of 8.44 on the synthetic data and 8.65 on MNIST, while achieving comparable accuracy and improved (lower) ECE relative to the other methods.}
    \vspace{6pt}
    \label{tab:class}
    \centering
    \begin{subtable}[t]{0.48\textwidth}
        \centering
        \includegraphics[width=\linewidth]{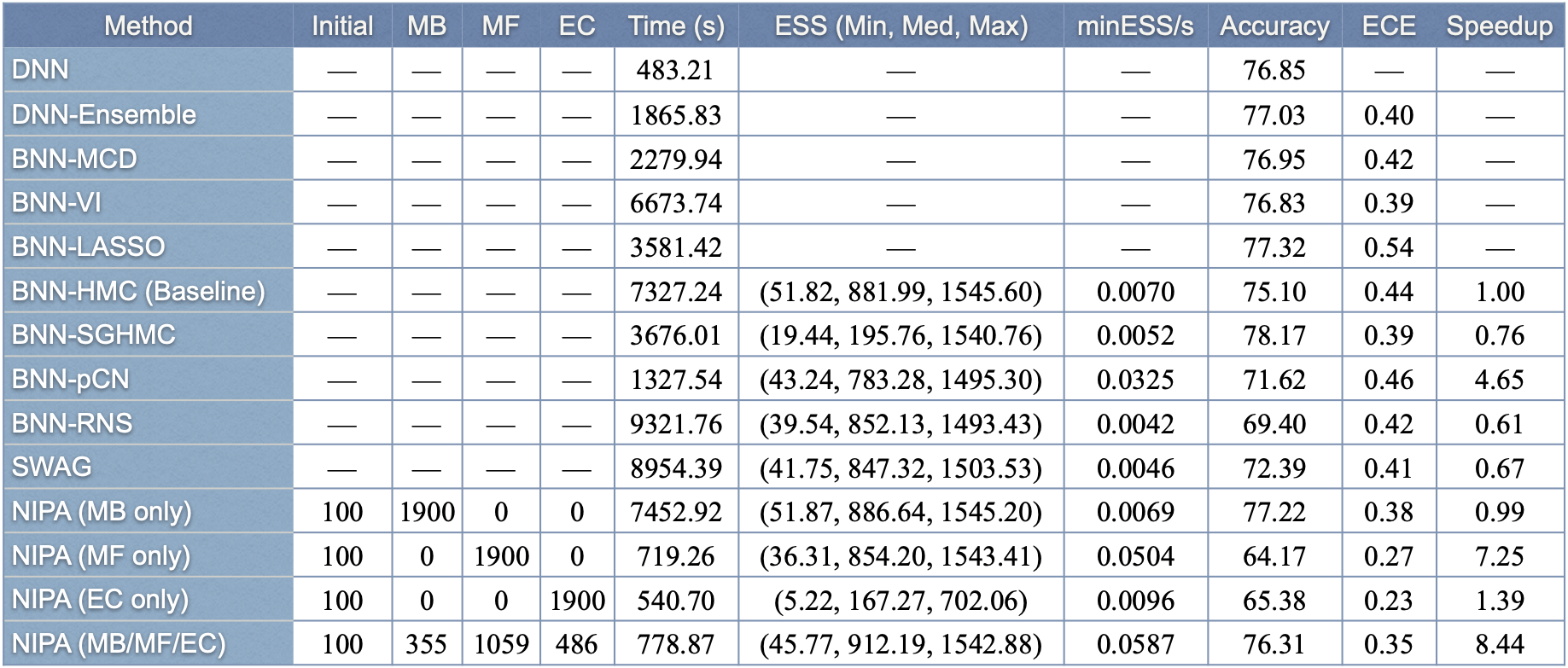}
        \vspace{-6pt}
        \caption{Synthetic Classification Problem. }
        \label{tab:nipa_sim}
    \end{subtable}
    \hspace{6pt}
    \begin{subtable}[t]{0.48\textwidth}
        \centering
        \includegraphics[width=\linewidth]{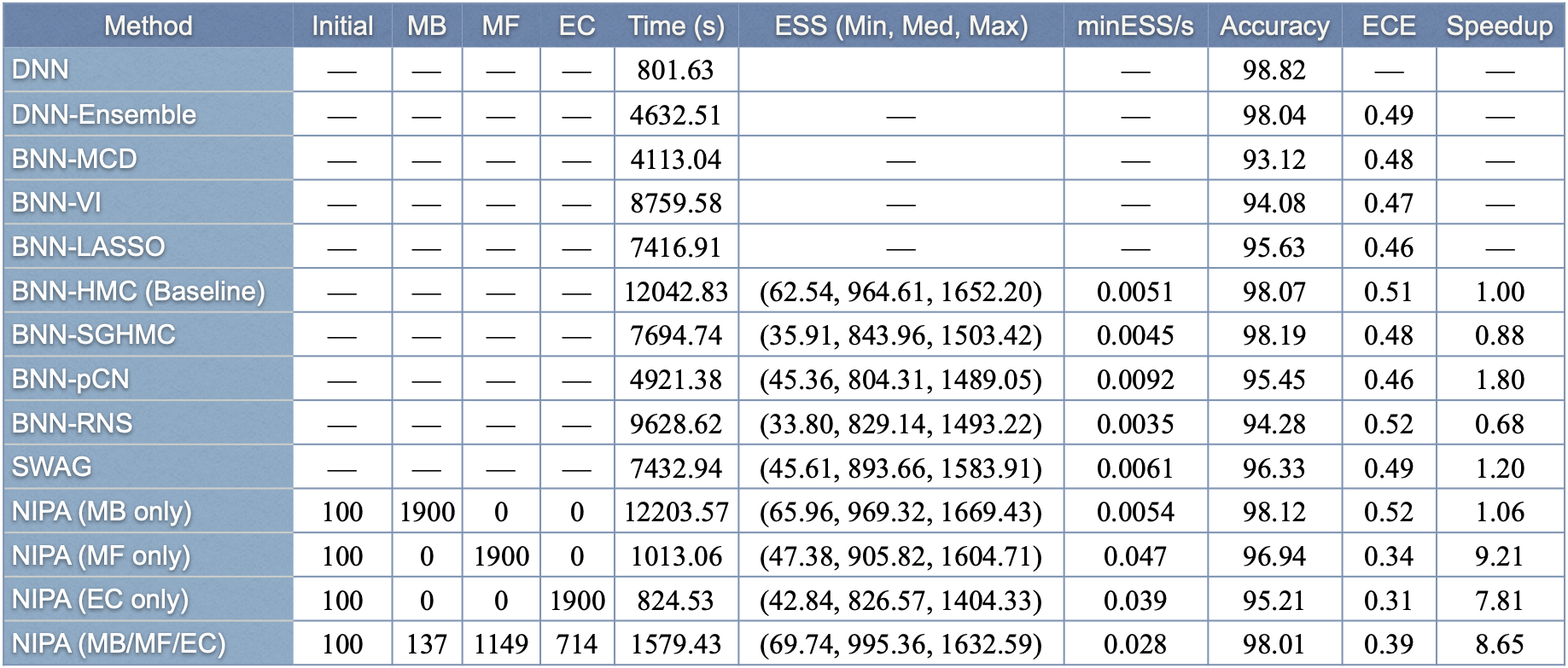}
        \vspace{-6pt}
        \caption{MNIST Odd vs. Even Digits.}
        \label{tab:nipa_mnist}
    \end{subtable}
    \label{tab:main}
    \vspace{-12pt}
\end{table}

\paragraph{Classification.}
Next, we construct a high-dimensional synthetic binary classification benchmark, also using a two-hidden-layer feedforward neural network with ReLU activations. Inputs are sampled as a standardized Gaussian design matrix $X\in\mathbb{R}^{N\times d}$. 
The network has hidden widths $h_1$ and $h_2$, with all weights and biases are concatenated into a parameter vector $\theta\in\mathbb{R}^D$, yielding a substantially high-dimensional parameter space.

We sample the ground-truth parameter vector $\theta^\star \sim \mathcal{N}\!\left(0, \Sigma \right)$, where $\Sigma$ is specified blockwise over weight and bias groups. Given $\theta^\star$, we compute the logits $\ell = f(X;\theta^\star) \in \mathbb{R}^N$ and sample binary labels according to the Bernoulli distribution:
\[
y_i \sim \mathrm{Bernoulli}\!\left(\operatorname{logit}^{-1}(\ell_i)\right), \qquad i = 1,\ldots,N.
\]
The resulting dataset has 20,000 observations, 512	features, and 147,841 model parameters.

We also evaluate our method on the MNIST dataset, which is commonly used as a benchmark for handwritten digit classification \citep{deng2012mnist}. Here, we focus on classifying odd versus even digits. This dataset has 70,000 observations, 784 features, and 217,538 model parameters

For each problem, we partition data into training and test sets, which are held fixed across all methods. 
The results on the test set for various models are provided in Table \ref{tab:class}. As we can see, NIPA is computationally more efficient than the BNN-HMC baseline, with comparable accuracy rate and improved (lower) ECE than other methods.

%% file: sections/discussion.tex
\section{Discussion}

We have introduced a neural-inspired Bayesian inference framework that aims to improve computational efficiency by mimicking how humans utilize multiple mechanisms to learn effectively from their environment and make decisions. This framework consists of three components that resemble model-based, model-free, and episodic control in the brain. Integrating these systems within a posterior approximation algorithm creates a general framework that includes many existing Bayesian computational methods as special cases. Additionally, we have shown that the resulting framework can improve the computational efficiency of Bayesian inference, particularly in deep learning settings where computational overhead has traditionally been prohibitive.

The specific settings of the algorithm presented in this paper are simple, intuitive, and effective in practice. Nonetheless, future work could refine these choices to improve generalizability and robustness across a broader range of models and datasets. For instance, the current version of NIPA relies on thresholds $t_1$ and $t_2$, which are treated as tuning parameters similar to the step size in HMC. However, selecting their optimal values is not trivial. Interestingly, in the brain, there is evidence that control is dynamically allocated among multiple mechanisms based on their relative uncertainty, providing a normative explanation for when flexible planning should dominate habitual efficiency \cite{DawNivDayan2005}. Following this idea, our gating approach could be extended to decide whether the more efficient MF or EC component should be used, based on their confidence in the estimated values. 

As mentioned earlier, many variations of this approach are possible, depending on the specific algorithms used to generate the initial sample pool, perform model-based sampling, construct surrogate functions for model-free sampling, and retrieve episodic memories. For example, instead of SGHMC, one could use a stochastic gradient optimization method and collect samples along the optimization trajectory (analogous to, but not identical to, SWAG). For model-based sampling, the preconditioned Crank–Nicolson (pCN) algorithm \cite{da2014} could be employed, as it is more effective at exploring distributions in the vicinity of their modes \cite{moslemi2024scaling}. Additionally, the gradient function used in the MB component could be approximated to enable faster implementations \cite{li2019neural}. For the surrogate model, simpler alternatives to DNNs, such as Extreme Learning Machines \cite{zhang16b}, could be adopted. Also, rather than relying on a single episode in the EC component, a $k$-nearest-neighbor model could be used to approximate the log-posterior using multiple nearby episodes.

Finally, although this paper focuses on applications in Bayesian deep learning, the proposed method can be readily extended to other computationally intensive Bayesian settings, such as Gaussian process models and Bayesian inverse problems for mechanistic models. While these approaches offer flexible and principled frameworks for modeling complex data, their high computational cost has often limited broader adoption.

%% file: sections/acknowledgment.tex
\section*{Acknowledgment}
We would like to thank Dr. Norbert Fortin (Neurobiology, UCI) and Dr. Aaron Bornstein (Cognitive Sciences, UCI) for insightful discussions and guidance on the neuroscience concepts motivating this work.

%% file: references/References.bib
@inproceedings{blundell2015weight,
  author={Blundell, Charles and Cornebise, Julien and Kavukcuoglu, Koray and Wierstra, Daan},
  title={Weight uncertainty in neural network},
  booktitle={International conference on machine learning},
  pages={1613--1622},
  year={2015},
  organization={PMLR}
}

@inproceedings{graves2011practical, author = {Graves, Alex}, title = {Practical variational inference for neural networks}, year = {2011}, isbn = {9781618395993}, publisher = {Curran Associates Inc.}, address = {Red Hook, NY, USA}, booktitle = {Proceedings of the 24th International Conference on Neural Information Processing Systems}, pages = {2348–2356}, numpages = {9}, location = {Granada, Spain}, series = {NIPS'11} }

@inproceedings{rezende2014stochastic,
    author = {Rezende, Danilo Jimenez and Mohamed, Shakir and Wierstra, Daan},
    title = {Stochastic backpropagation and approximate inference in deep generative models},
    booktitle = {International conference on machine learning},
    organization = {PMLR},
    pages = {1278--1286},
    year = {2014}
}

@article{maddox2019simple,
    author = {Maddox, Wesley J and Izmailov, Pavel and Garipov, Timur and Vetrov, Dmitry P and Wilson, Andrew Gordon},
    title = {{A simple baseline for Bayesian uncertainty in deep learning}},
    journal = {Advances in neural information processing systems},
    volume = {32},
    year = {2019}
}

@inproceedings{gal2016dropout,
    author = {Gal, Yarin and Ghahramani, Zoubin},
    title = {Dropout as a Bayesian approximation: Representing model uncertainty in deep learning},
    booktitle = {International conference on machine learning},
    pages = {1050--1059},
    organization = {PMLR},
    year = {2016}
}

@inproceedings{izmailov2018,
  author       = {Pavel Izmailov and
                  Dmitrii Podoprikhin and
                  Timur Garipov and
                  Dmitry P. Vetrov and
                  Andrew Gordon Wilson},
  editor       = {Amir Globerson and
                  Ricardo Silva},
  title        = {Averaging Weights Leads to Wider Optima and Better Generalization},
  booktitle    = {Proceedings of the Thirty-Fourth Conference on Uncertainty in Artificial
                  Intelligence, {UAI} 2018, Monterey, California, USA, August 6-10,
                  2018},
  pages        = {876--885},
  publisher    = {{AUAI} Press},
  year         = {2018}
}

@article{li2019neural,
    author = {Li, Lingge and Holbrook, Andrew and Shahbaba, Babak and Baldi, Pierre},
    title = {{Neural network gradient Hamiltonian Monte Carlo}},
    journal = {Computational Statistics},
    volume = {34},
    pages = {281--299},
    year = {2019},
    publisher = {Springer}
}

@article{jospin2022,
    author = {Jospin, Laurent Valentin and Laga, Hamid and Boussaid, Farid and Buntine, Wray and Bennamoun, Mohammed},
    title = {Hands-on {Bayesian} neural networks---A tutorial for deep learning users},
    journal = {IEEE Computational Intelligence Magazine},
    volume = {17},
    number = {2},
    pages = {29--48},
    year = {2022},
    publisher = {IEEE}
}

@article{Cleary2021,
    author = {Cleary, Emmet and Garbuno-Inigo, Alfredo and Lan, Shiwei and Schneider, Tapio and Stuart, Andrew M.},
    title = {Calibrate, emulate, sample},
    journal = {Journal of Computational Physics},
    volume = {424},
    pages = {109716},
    year = {2021},
    doi = {10.1016/j.jcp.2020.109716},
    publisher = {Elsevier BV}
}

@misc{year_prediction_msd_203,
  author       = {Bertin-Mahieux, Thierry},
  title        = {{Year Prediction MSD}},
  year         = {2011},
  howpublished = {UCI Machine Learning Repository},
  note         = {{DOI}: https://doi.org/10.24432/C50K61}
}

@article{deng2012mnist,
  title={The mnist database of handwritten digit images for machine learning research},
  author={Deng, Li},
  journal={IEEE Signal Processing Magazine},
  volume={29},
  number={6},
  pages={141--142},
  year={2012},
  publisher={IEEE}
}

@inproceedings{chuan2017,
    author = {Guo, Chuan and Pleiss, Geoff and Sun, Yu and Weinberger, Kilian Q},
    title = {On Calibration of Modern Neural Networks},
    booktitle = {Proceedings of the 34th International Conference on Machine Learning},
    pages = {1321--1330},
    year = {2017},
    publisher = {PMLR},
    month = {August}
}

@inproceedings{nixon2019,
    author = {Nixon, Jeremy and Dusenberry, Michael W and Zhang, Linchuan and Jerfel, Ghassen and Tran, Dustin},
    title = {Measuring Calibration in Deep Learning},
    booktitle = {Proceedings of the IEEE/CVF Conference on Computer Vision and Pattern Recognition (CVPR) Workshops},
    year = {2019}
}

@article{mackay1992,
    author = {MacKay, David JC},
    title = {A practical Bayesian framework for backpropagation networks},
    journal = {Neural Computation},
    volume = {4},
    number = {3},
    pages = {448--472},
    year = {1992},
    publisher = {MIT Press}
}

@book{da2014,
    author = {Da Prato, Giuseppe and Zabczyk, Jerzy},
    title = {Stochastic equations in infinite dimensions},
    publisher = {Cambridge University Press},
    year = {2014}
}

@inproceedings{chen2014,
    author = {Chen, Tianqi and Fox, Emily and Guestrin, Carlos},
    title = {Stochastic gradient {Hamiltonian Monte Carlo}},
    booktitle = {International Conference on Machine Learning},
    pages = {1683--1691},
    year = {2014}
}

@article{liu2020,
    author = {Liu, Haitao and Ong, Yew-Soon and Shen, Xiaobo and Cai, Jianfei},
    title = {When Gaussian process meets big data: A review of scalable GPs},
    journal = {IEEE Transactions on Neural Networks and Learning Systems},
    volume = {31},
    number = {11},
    pages = {4405--4423},
    publisher = {IEEE},
    year = {2020}
}

@article{bonilla2007,
    author = {Bonilla, Edwin V and Chai, Kian and Williams, Christopher},
    title = {Multi-task Gaussian process prediction},
    journal = {Advances in Neural Information Processing Systems},
    volume = {20},
    year = {2007}
}

@article{gardner2018,
    author = {Gardner, Jacob and Pleiss, Geoff and Weinberger, Kilian Q and Bindel, David and Wilson, Andrew G},
    title = {{GPyTorch}: Blackbox matrix-matrix Gaussian process inference with {GPU} acceleration},
    journal = {Advances in Neural Information Processing Systems},
    volume = {31},
    year = {2018}
}

@inproceedings{seeger2003,
    author = {Seeger, Matthias W. and Williams, Christopher K. I. and Lawrence, Neil D.},
    title = {Fast forward selection to speed up sparse Gaussian process regression},
    booktitle = {International Workshop on Artificial Intelligence and Statistics},
    pages = {254--261},
    year = {2003},
    organization = {PMLR}
}

@article{lakshminarayanan2017,
  title={Simple and scalable predictive uncertainty estimation using deep ensembles},
  author={Lakshminarayanan, Balaji and Pritzel, Alexander and Blundell, Charles},
  journal={Advances in neural information processing systems},
  volume={30},
  year={2017}
}

@InProceedings{Campbell2024,
  title     = {Coreset Markov chain Monte Carlo},
  author    = {Chen, Naitong and Campbell, Trevor},
  booktitle = {Proceedings of The 27th International Conference on Artificial Intelligence and Statistics},
  editor    = {Dasgupta, Sanjoy and Mandt, Stephan and Li, Yingzhen},
  series    = {Proceedings of Machine Learning Research},
  volume    = {238},
  pages     = {4438--4446},
  year      = {2024},
  publisher = {PMLR},
  month     = {02--04 May}
}

@article{Marzouk2014,
  author     = "Li, J. and Marzouk, Y. M.",
  journal    = "SIAM Journal on Scientific Computing",
  number     = "3",
  pages      = "A1163--A1186",
  title      = "Adaptive construction of surrogates for the Bayesian solution of inverse problems",
  volume     = "36",
  year       = "2014"
}

@incollection{geyer2011introduction,
  title={Introduction to Markov Chain Monte Carlo},
  author={Geyer, Charles J.},
  booktitle={Handbook of Markov Chain Monte Carlo},
  editor={Brooks, Steve and Gelman, Andrew and Jones, Galin L. and Meng, Xiao-Li},
  pages={3--48},
  year={2011},
  publisher={Chapman \& Hall/CRC}
}

@inproceedings{naeini2015bayesian,
  title={Obtaining Well Calibrated Probabilities Using Bayesian Binning},
  author={Naeini, Mahdi Pakdaman and Cooper, Gregory F. and Hauskrecht, Milos},
  booktitle={Proceedings of the Twenty-Ninth AAAI Conference on Artificial Intelligence},
  year={2015}
}

@InProceedings{guo17calibration,
  title     = {On Calibration of Modern Neural Networks},
  author    = {Chuan Guo and Geoff Pleiss and Yu Sun and Kilian Q. Weinberger},
  booktitle = {Proceedings of the 34th International Conference on Machine Learning},
  editor    = {Doina Precup and Yee Whye Teh},
  series    = {Proceedings of Machine Learning Research},
  volume    = {70},
  pages     = {1321--1330},
  year      = {2017},
  publisher = {PMLR},
  url       = {https://proceedings.mlr.press/v70/guo17a.html}
}

@article{Coullon2023Efficient,
  title        = {Efficient and generalizable tuning strategies for stochastic gradient MCMC},
  author       = {Coullon, Jeremie and South, Leah F. and Nemeth, Christopher},
  journal      = {Statistics and Computing},
  volume       = {33},
  number       = {3},
  pages        = {66},
  year         = {2023},
  doi          = {10.1007/s11222-023-10233-3},
  url          = {https://doi.org/10.1007/s11222-023-10233-3}
}

@inproceedings{alexos22,
  title     = {Structured Stochastic Gradient MCMC},
  author    = {Alexos, Antonios and Boyd, Alex J. and Mandt, Stephan},
  booktitle = {Proceedings of the 39th International Conference on Machine Learning},
  pages     = {414--434},
  year      = {2022},
  editor    = {Chaudhuri, Kamalika and Jegelka, Stefanie and Song, Le and Szepesvari, Csaba and Niu, Gang and Sabato, Sivan},
  volume    = {162},
  series    = {Proceedings of Machine Learning Research},
  publisher = {PMLR},
  url       = {https://proceedings.mlr.press/v162/alexos22a.html}
}

@article{nemeth2021stochastic,
  title   = {Stochastic Gradient Markov Chain Monte Carlo},
  author  = {Nemeth, Christopher and Fearnhead, Paul},
  journal = {Journal of the American Statistical Association},
  volume  = {116},
  number  = {533},
  pages   = {433--450},
  year    = {2021},
  publisher = {Taylor \& Francis}
}

@inproceedings{kingma2014VAE,
  author    = {Kingma, Diederik P. and Welling, Max},
  title     = {Auto-Encoding Variational Bayes},
  booktitle = {Proceedings of the International Conference on Learning Representations (ICLR)},
  year      = {2014}
}

@inproceedings{meeds14,
  author    = {Meeds, Edward and Welling, Max},
  title     = {GPS-ABC: Gaussian Process Surrogate Approximate Bayesian Computation},
  booktitle = {Proceedings of the Thirtieth Conference on Uncertainty in Artificial Intelligence (UAI)},
  year      = {2014},
  pages     = {593--602},
  address   = {Quebec City, QC, Canada},
  publisher = {AUAI Press},
  note      = {arXiv:1401.2838},
  url       = {https://arxiv.org/abs/1401.2838}
}

@article{
moslemi2024scaling,
title={Scaling Up Bayesian Neural Networks with Neural Networks},
author={Zahra Moslemi and Yang Meng and Shiwei Lan and Babak Shahbaba},
journal={Transactions on Machine Learning Research},
issn={2835-8856},
year={2024},
url={https://openreview.net/forum?id=cD209UgOX7},
note={}
}

@article{Blei17,
  title   = {Variational Inference: A Review for Statisticians},
  author  = {Blei, David M. and Kucukelbir, Alp and McAuliffe, Jon D.},
  journal = {Journal of the American Statistical Association},
  volume  = {112},
  number  = {518},
  pages   = {859--877},
  year    = {2017}
}

@article{DawDayan2014,
  author       = {Daw, Nathaniel D. and Dayan, Peter},
  title        = {The algorithmic anatomy of model-based evaluation},
  journal      = {Philosophical Transactions of the Royal Society B: Biological Sciences},
  year         = {2014},
  volume       = {369},
  number       = {1655},
  pages        = {20130478},
  doi          = {10.1098/rstb.2013.0478},
}

@article{DawNivDayan2005,
  author       = {Daw, Nathaniel D. and Niv, Yael and Dayan, Peter},
  title        = {Uncertainty-based competition between prefrontal and dorsolateral striatal systems for behavioral control},
  journal      = {Nature Neuroscience},
  year         = {2005},
  volume       = {8},
  number       = {12},
  pages        = {1704--1711},
  doi          = {10.1038/nn1560},
}

@article{GershmanDaw2017,
  author       = {Gershman, Samuel J. and Daw, Nathaniel D.},
  title        = {Reinforcement Learning and Episodic Memory in Humans and Animals: An Integrative Framework},
  journal      = {Annual Review of Psychology},
  year         = {2017},
  volume       = {68},
  pages        = {101--128},
  doi          = {10.1146/annurev-psych-122414-033625},
}

@inproceedings{LengyelDayan2007,
  author       = {Lengyel, M{\'a}t{\'e} and Dayan, Peter},
  title        = {Hippocampal Contributions to Control: The Third Way},
  booktitle    = {Advances in Neural Information Processing Systems 20},
  series       = {NeurIPS ’07},
  editor       = {Platt, J. C. and Koller, D. and Singer, Y. and Roweis, S. T.},
  pages        = {889--896},
  year         = {2007},
  publisher    = {Curran Associates, Inc.},
  address      = {Red Hook, NY, USA},
}

@article{Zhang2017a,
author = {Zhang, C. and Shahbaba, B. and Zhao, H.},
doi = {10.1007/s11222-016-9699-1},
issn = {15731375},
journal = {Statistics and Computing},
number = {6},
title = {{Hamiltonian Monte Carlo acceleration using surrogate functions with random bases}},
volume = {27},
year = {2017}
}

@article{Li2019,
  title   = {Neural Network Gradient Hamiltonian Monte Carlo},
  author  = {Lingge Li and Andrew Holbrook and Babak Shahbaba and Pierre Baldi},
  journal = {Computational Statistics},
  volume  = {34},
  number  = {1},
  pages   = {281--299},
  year    = {2019},
  doi     = {10.1007/s00180-018-00861-x},
  eprint  = {arXiv:1711.05307},
  archivePrefix = {arXiv},
  primaryClass  = {stat.ME}
}

@article{Lan2022Siam,
author = {Lan, Shiwei and Li, Shuyi and Shahbaba, Babak},
title = {Scaling Up Bayesian Uncertainty Quantification for Inverse Problems Using Deep Neural Networks},
journal = {SIAM/ASA Journal on Uncertainty Quantification},
volume = {10},
number = {4},
pages = {1684-1713},
year = {2022},
doi = {10.1137/21M1439456}
}

@inproceedings{chen2014stochastic,
  title={Stochastic gradient hamiltonian monte carlo},
  author={Chen, Tianqi and Fox, Emily and Guestrin, Carlos},
  booktitle={International Conference on Machine Learning},
  pages={1683--1691},
  year={2014}
}

@article{wainwright08,
	Author = {M. Wainwright and M. Jordan},
	Journal = {Foundations and Trends in Machine Learning},
	Number = {1-2},
	Pages = {1--305},
	Title = {{Graphical models, exponential families, and variational inference}},
	Volume = {1},
	Year = {2008}}

@inproceedings{salimans15,
	Author = {Tim Salimans and Diederik Kingma and Max Welling},
	Booktitle = {Proceedings of the 32nd International Conference on Machine Learning},
	Editor = {Francis Bach and David Blei},
	Pages = {1218--1226},
	Publisher = {PMLR},
	Title = {Markov Chain Monte Carlo and Variational Inference: Bridging the Gap},
	Volume = {37},
	Year = {2015}}

@inproceedings{betancourt15,
	Author = {Michael Betancourt},
	Booktitle = {Proceedings of the 32nd International Conference on Machine Learning, {ICML} 2015, Lille, France, 6-11 July 2015},
	Pages = {533--540},
	Title = {The Fundamental Incompatibility of Scalable Hamiltonian Monte Carlo and Naive Data Subsampling},
	Year = {2015}}

@article{lan15,
	Author = {Lan, S. and Palacios, J.~A. and Karcher, M. and Minin, V. and Shahbaba, B},
	Journal = {Bioinformatics},
	Number = {20},
	Pages = {3282-3289},
	Title = {An Efficient Bayesian Inference Framework for Coalescent-Based Nonparametric Phylodynamics},
	Volume = {31},
	Year = {2015}}

@article{chen14,
	Author = {Chen, T. and Fox, E.~B. and Guestrin, C.},
	Journal = {Preprint},
	Title = {Stochastic Gradient Hamiltonian Monte Carlo},
	Year = 2014}

@article{rasmussen03,
	Author = {Rasmussen, C.~E.},
	Journal = {Bayesian Statistics},
	Pages = {651-659},
	Title = {Gaussian Processes to Speed up Hybrid Monte Carlo for Expensive Bayesian Integrals},
	Year = 2003}

@article{zhang16c,
	Author = {Zhang, C. and Shahbaba, B. and Zhao, H.},
	Journal = {Submitted},
	Title = {{Variational Hamiltonian Monte Carlo via Score Matching}},
	Url = {https://arxiv.org/abs/1602.02219},
	Year = {2016}}

@article{zhang16b,
	Author = {Zhang, C. and Shahbaba, B. and Zhao, H.},
	Journal = {Statistics and Computing (to appear)},
	Title = {{Hamiltonian Monte Carlo Acceleration Using Neural Network Surrogate functions}},
	Url = {http://arxiv.org/abs/1506.05555},
	Year = {2015}}

@conference{AhnShahbabaWelling14,
	Author = {S. Ahn and B. Shahbaba and M. Welling},
	Booktitle = {International Conference on Machine Learning},
	Title = {{Distributed Stochastic Gradient MCMC}},
	Year = {2014}}

@inproceedings{hoffmann10,
	Author = {Hoffman, Matthew D. and Blei, David M. and Bach, Francis R.},
	Booktitle = {NIPS},
	Editor = {Lafferty, John D. and Williams, Christopher K. I. and Shawe-Taylor, John and Zemel, Richard S. and Culotta, Aron},
	Pages = {856-864},
	Publisher = {Curran Associates, Inc.},
	Title = {Online Learning for Latent Dirichlet Allocation.},
	Year = 2010}

@inproceedings{lanAAAI14,
	Author = {Lan, S. and Streets, J. and Shahbaba, B.},
	Booktitle = {Proceedings of the Twenty-Eighth AAAI Conference on Artificial Intelligence},
	Title = {Wormhole Hamiltonian Monte Carlo},
	Year = {2014}}

@conference{lanICML14,
	Author = {Lan, S. and Zhou, B. and Shahbaba, B.},
	Booktitle = {Proceedings of the 31th International Conference on Machine Learning (ICML)},
	Title = {{Spherical Hamiltonian Monte Carlo for constrained target distributions}},
	Year = {(2014)}}

@incollection{calderhead12,
  title     = {Sparse Approximate Manifolds for Differential Geometric MCMC},
  author    = {Calderhead, Ben and Sustik, Mátyás A.},
  booktitle = {Advances in Neural Information Processing Systems 25},
  pages     = {2888--2896},
  year      = {2012},
  publisher = {Curran Associates, Inc.},
}

@inproceedings{ZhaSut2011a,
	Author = {Y. Zhang and C. Sutton},
	Booktitle = {Advances In Neural Information Processing Systems},
	Title = {Quasi-{N}ewton {M}arkov Chain {M}onte {C}arlo},
	Year = {2011}}

@article{cappe08,
	Author = {Capp\'{e}, Olivier and Douc, Randal and Guillin, Arnaud and Marin, Jean-Michel and Robert, Christian P.},
	Journal = {Statistics and Computing},
	Number = {4},
	Pages = {447--459},
	Title = {Adaptive importance sampling in general mixture classes},
	Volume = {18},
	Year = {2008}}

@phdthesis{Beal03,
	Address = {London, UK},
	Author = {Beal, M. J.},
	Institution = {Gatsby Computational Neuroscience Unit, University College London},
	School = {University College London},
	Title = {Variational Algorithms for Approximate {Bayesian} Inference},
	Year = {2003}}

@inproceedings{KuriharaWellingVlassis06,
	Author = {K. Kurihara and M. Welling and N. Vlassis},
	Booktitle = {Advances of Neural Information Processing Systems -- NIPS},
	Title = {Accelerated Variational {Dirichlet} Process mixtures},
	Volume = {19},
	Year = {2006}}

@article{MollerPettittBerthelsenReeves06,
	Author = {J. M{\o}ller and A. Pettitt and K. Berthelsen and R. Reeves},
	Journal = {Biometrica},
	Note = {to appear},
	Title = {An efficient {M}arkov chain {M}onte {C}arlo method for distributions with intractable normalisation constants.},
	Volume = {93},
	Year = {2006}}

@article{ProppWilson96,
  author    = {Propp, James G. and Wilson, David B.},
  title     = {Exact sampling with coupled Markov chains and applications to statistical mechanics},
  journal   = {Random Structures \& Algorithms},
  volume    = {9},
  pages     = {223--252},
  year      = {1996},
  doi       = {10.1002/(SICI)1098-2418(199608/09)9:1/2<223::AID-RSA14>3.0.CO;2-O},
}

@inproceedings{GelfandMaatenChenWelling10,
	Author = {A. Gelfand and L. van der Maaten and Y. Chen and M. Welling},
	Booktitle = {Advances in Neural Information Processing Systems 23},
	Pages = {694-702},
	Title = {On Herding and the Cycling Perceptron Theorem},
	Year = {2010}}

@inproceedings{wellingUAI09,
	Author = {Welling, M.},
	Booktitle = {Proc. of Intl. Conf. on Machine Learning},
	Title = {Herding Dynamic Weights to Learn},
	Year = 2009}

@article{AndrieuMoulines06,
	Author = {C. Andrieu and E. Moulines},
	Journal = {Annals of Applied Probability},
	Number = {3},
	Pages = {1462-1505},
	Title = {On the ergodicity properties of some adaptive MCMC algorithms},
	Volume = {16},
	Year = {2006}}

@article{douc11,
	Author = {Randal, D. and Christian R. P.},
	Journal = {Annals of Statistics},
	Number = {1},
	Pages = {261-277},
	Publisher = {Institute of Mathematical Statistics},
	Title = {A vanilla Rao-Blackwellization of Metropolis-Hastings algorithms},
	Volume = {39},
	Year = {2011}}

@article{douc07,
	Author = {Randal, R. and Arnaud, G. and Jean-Michel, M. and Christian, R. P.},
	Journal = {ESAIM: Probability and Statistics},
	Pages = {427-447},
	Publisher = {Cambridge University Press},
	Title = {Minimum variance importance sampling via Population Monte Carlo},
	Volume = {11},
	Year = {2007}}

@article{gilks98,
	Author = {W. R. Gilks and G. O. Roberts and S. K. Sahu},
	Journal = {Journal of the American Statistical Association},
	Number = {443},
	Pages = {1045--1054},
	Title = {Adaptive Markov chain Monte Carlo through regeneration},
	Volume = {93},
	Year = {1998}}

@article{mykland95,
	Author = {Mykland, P. and Tierney, L. and Yu, B.},
	Journal = {Journal of the American Statistical Association},
	Number = {429},
	Pages = {233--241},
	Title = {Regeneration in {M}arkov {C}hain {S}amplers},
	Volume = {90},
	Year = {1995}}

@inproceedings{Welling11,
	Author = {M. Welling and Y. W. Teh},
	Booktitle = {Proceedings of the International Conference on Machine Learning},
	Title = {{B}ayesian Learning via Stochastic Gradient {L}angevin Dynamics},
	Year = {2011}}

@techreport{warnes01,
	Author = {Warnes, G. R.},
	Institution = {University of Washington},
	Number = {Technical Report No. 395},
	Title = {{The normal kernel coupler: An adaptive Markov Chain Monte Carlo method for efficiently sampling from multi-modal distributions}},
	Year = {2001}}

@article{craiu09,
	Author = {Craiu, R. V. and R., Jeffrey and Y., Chao},
	Journal = {Journal of the American Statistical Association},
	Number = {488},
	Pages = {1454-1466},
	Title = {Learn From Thy Neighbor: Parallel-Chain and Regional Adaptive MCMC},
	Volume = {104},
	Year = {2009}}

@conference{ahn13,
	Author = {Ahn, S. and Chen, Y. and Welling, M.},
	Booktitle = {Proceedings of the 16th International Conference on Artificial Intelligence and Statistics (AI Stat)},
	Title = {{Distributed and adaptive darting Monte Carlo through regenerations}},
	Year = {2013}}

@conference{ahn12,
	Author = {Ahn, Sungjin and Korattikara, Anoop and Welling, Max},
	Booktitle = {Proceedings of the 29th International Conference on Machine Learning (ICML)},
	Pages = {1591--1598},
	Title = {Bayesian Posterior Sampling via Stochastic Gradient Fisher Scoring},
	Year = {2012}}

@article{ahmadian11,
	Author = {Ahmadian, Y. and Pillow, J. W. and Paninski, L.},
	Journal = {Neural Computation},
	Number = {1},
	Pages = {46-96},
	Title = {Efficient {Markov Chain Monte Carlo} Methods for Decoding Neural Spike Trains},
	Volume = {23},
	Year = {2011}}

@misc{hoffman11,
	Author = {Hoffman, M. and Gelman, A.},
	Howpublished = {arxiv.org/abs/1111.4246},
	Title = {{The No-U-Turn Sampler: Adaptively Setting Path Lengths in Hamiltonian Monte Carlo}},
	Year = {2011}}

@article{brockwell06,
	Author = {Brockwell, A. E.},
	Journal = {Journal of Computational and Graphical Statistics},
	Pages = {246--261},
	Title = {Parallel Markov Chain Monte Carlo Simulation by {Pre-Fetching}},
	Year = {2006}}

@inproceedings{freitas01,
	Address = {San Francisco, CA, USA},
	Author = {de Freitas, N. and H{\o}jen-S{\o}rensen, P. and Jordan, M. and Stuart, R.},
	Booktitle = {Proceedings of the 17th Conference in Uncertainty in Artificial Intelligence},
	Isbn = {1-55860-800-1},
	Pages = {120--127},
	Publisher = {Morgan Kaufmann Publishers Inc.},
	Series = {UAI '01},
	Title = {{Variational MCMC}},
	Year = {2001}}

@conference{WellingTeh11,
	Author = {M. Welling and Y.W. Teh},
	Booktitle = {Proceedings of the 28th International Conference on Machine Learning (ICML)},
	Pages = {681-688},
	Title = {Bayesian Learning via Stochastic Gradient Langevin Dynamics},
	Year = {2011}}

@article{shahbabaSplitHMC,
	Author = {Shahbaba, B. and Lan, S. and Johnson, W.O. and Neal, R.M.},
	Journal = {Statistics and Computing},
	Number = {3},
	Pages = {339-349},
	Title = {{Split Hamiltonian Monte Carlo}},
	Volume = {24},
	Year = {2014}}

@article{lanLMC15,
	Author = {Lan, S. and Stathopoulos, V. and Shahbaba, B. and Girolami, M.},
	Howpublished = {arxiv.org/abs/1211.3759},
	Journal = {Journal of Computational and Graphical Statistics},
	Number = {2},
	Pages = {357-378},
	Title = {{Markov Chain Monte Carlo from Lagrangian Dynamics}},
	Volume = {24},
	Year = {2015}}

@article{PMCMC,
	Author = {C. J. Geyer},
	Journal = {Statistical Science},
	Number = {4},
	Pages = {473-483},
	Title = {{Practical Markov Chain Monte Carlo}},
	Volume = {7},
	Year = {1992}}

@article{roberts97,
	Author = {Roberts, G. O. and Sahu, S. K.},
	Journal = {Journal of the Royal Statistical Society, Series B},
	Pages = {291-317},
	Title = {{Updating Schemes, Correlation Structure, Blocking and Parameterisation for the Gibbs Sampler}},
	Volume = {59},
	Year = {1997}}

@article{park08,
	Author = {Park, T. and Casella, G.},
	Journal = {Journal of the American Statistical Association},
	Month = {June},
	Number = {482},
	Pages = {681--686},
	Publisher = {American Statistical Association},
	Title = {{The Bayesian Lasso}},
	Volume = {103},
	Year = {2008}}

@article{girolami11,
	Author = {Girolami, M. and Calderhead, B.},
	Journal = {Journal of the Royal Statistical Society, Series B},
	Number = {2},
	Pages = {123--214},
	Title = {{Riemann manifold Langevin and Hamiltonian Monte Carlo methods}},
	Volume = {(with discussion) 73},
	Year = {2011}}

@article{beskos11,
	Author = {A. Beskos and F. J. Pinski and J. M. Sanz-Serna and A. M. Stuart},
	Journal = {Stochastic Processes and their Applications},
	Pages = {2201-2230},
	Title = {{Hybrid Monte-Carlo on Hilbert spaces}},
	Volume = {121},
	Year = {2011}}

@article{geyer92,
	Author = {Geyer, C. J.},
	Journal = {Statistical Science},
	Number = {4},
	Pages = {473-483},
	Title = {{Practical Markov Chain Monte Carlo}},
	Volume = {7},
	Year = {1992}}

@incollection{neal11,
	Author = {Neal, R.M.},
	Booktitle = {Handbook of Markov Chain Monte Carlo},
	Editor = {Brooks, S. and Gelman, A. and Jones, G. and Meng, X. L.},
	Pages = {113-162},
	Publisher = {Chapman and Hall/CRC},
	Title = {{MCMC using Hamiltonian dynamics}},
	Year = {2011}}

@article{duane87,
	Author = {Duane, S. and Kennedy, A. D. and Pendleton, B J. and Roweth, D.},
	Journal = {Physics Letters B},
	Number = {2},
	Pages = {216 - 222},
	Title = {{Hybrid Monte Carlo}},
	Volume = {195},
	Year = {1987}}

@article{metropolis59,
	Author = {Metropolis, N. and Rosenbluth, A. W. and Rosenbluth, M. N. and Teller, A. H. and Teller, E.},
	Journal = {The Journal of Chemical Physics},
	Number = {6},
	Pages = {1087--1092},
	Title = {{Equation of State Calculations by Fast Computing Machines}},
	Volume = {21},
	Year = {1953}}

@inproceedings{Jordan99,
	Author = {Jordan, M. I. and Ghahramani, Z. and Jaakkola, T. S. and Saul, L. K.},
	Booktitle = {Machine Learning},
	Pages = {183--233},
	Publisher = {MIT Press},
	Title = {An Introduction to Variational Methods for Graphical Methods},
	Year = {1999}}

@techreport{neal05,
	Author = {Neal, R.M.},
	Institution = {Department of Statistics, University of Toronto},
	Number = {0506},
	Title = {The short-cut Metropolis method},
	Year = {2005}}

@book{neal96,
	Author = {Neal, R.M.},
	Publisher = {Lecture Notes in Statistics No. 118, New York: Springer-Verlag},
	Title = {Bayesian Learning for Neural Networks},
	Year = {1996}}

@article{neal96a,
	Author = {Neal, R.M.},
	Journal = {Statistics and Computing},
	Number = {4},
	Pages = {353},
	Title = {Sampling from multimodal distributions using tempered transitions},
	Volume = {6},
	Year = {1996}}

@book{neal93,
	Author = {Neal, R.M.},
	Publisher = {Technical Report CRG-TR-93-1, Department of Computer Science, University of Toronto},
	Title = {Probabilistic Inference Using {Markov} Chain {Monte Carlo} Methods},
	Year = {1993}}

@article{neal03,
	Author = {Neal, R.M.},
	Journal = {Annals of Statistics},
	Number = {3},
	Pages = {705-767},
	Title = {Slice sampling},
	Volume = {31},
	Year = {2003}}

@book{liu01,
  author    = {Liu, J. S.},
  title     = {Monte Carlo Strategies in Scientific Computing},
  publisher = {Springer},
  year      = {2001}
}

@article{zhang15,
  author  = {Zhang, C. and Shahbaba, B. and Zhao, H.},
  title   = {Precomputing Strategy for Hamiltonian Monte Carlo Method Based on Regularity in Parameter Space},
  journal = {arXiv preprint arXiv:1504.01418},
  year    = {2015}
}

@article{strathmann15,
  author  = {Strathmann, H. and Sejdinovic, D. and Livingstone, S. and Szabo, Z. and Gretton, A.},
  title   = {Gradient-free Hamiltonian Monte Carlo with Efficient Kernel Exponential Families},
  journal = {arXiv preprint arXiv:1506.02564},
  year    = {2015}
}

@incollection{Ding14,
title = {Bayesian Sampling Using Stochastic Gradient Thermostats},
author = {Ding, Nan and Fang, Youhan and Babbush, Ryan and Chen, Changyou and Skeel, Robert D and Neven, Hartmut},
booktitle = {Advances in Neural Information Processing Systems 27},
editor = {Z. Ghahramani and M. Welling and C. Cortes and N. D. Lawrence and K. Q. Weinberger},
pages = {3203--3211},
year = {2014},
publisher = {Curran Associates, Inc.}
}
